\documentclass{article} % For LaTeX2e
\usepackage{iclr2016_conference,times}
\usepackage{hyperref}
\usepackage{url}
\usepackage{graphicx}
\usepackage{array}
\newcolumntype{P}[1]{>{\centering\arraybackslash}p{#1}}
\graphicspath{ {images/} }

\title{Dynamic Adaptive Network Intelligence}

\author{Richard Searle \& Megan Bingham-Walker \\
Eccentric Data Limited \\
Botanic House, 100 Hills Road \\
Cambridge, UK \\
\texttt{\{richard,megan\}@eccentricdata.com} \\
}

\begin{document}

\maketitle

\begin{abstract}
Accurate representational learning of both the explicit and implicit relationships within data is critical to the ability of machines to perform more complex and abstract reasoning tasks. We describe the efficient weakly supervised learning of such inferences by our Dynamic Adaptive Network Intelligence (DANI) model. We report state-of-the-art results for DANI over question answering tasks in the bAbI dataset that have proved difficult for contemporary approaches to learning representation \citep{weston2015towards}.       
\end{abstract}

\section{Introduction}
 
The Facebook bAbI dataset was proposed by \citet{weston2015towards} to demonstrate the efficiency of new algorithms for machine reading and comprehension.  Despite some success with adding a memory component to deep learning models for question answering, tasks requiring inference and reasoning remain difficult to solve in the absence of sufficient training data and strong supervision \citep{weston2015towards,weston2014memory,sukhbaatar2015end,sukhbaatar2015weakly,kumar2015ask}.  

We set out to learn a new type of representational model, which we call Dynamic Adaptive Network Intelligence (DANI). This is a weakly supervised, network-based representation of the data with efficient model constraints to enable scalability. We use DANI to perform complex reasoning over the observed data to solve the specified bAbI tasks.  Preliminary results, reported in this paper demonstrate that DANI is able to learn a simple domain representation from significantly less training data and with less supervision than in systems disclosed in previous research. 

In presenting our early work in this field, we seek to demonstrate that accurate inferential reasoning can be achieved using a basic graphical technique, without the requirement for strict supervision or priming, and without incurring a decay in performance when scaling.

\section{Methods}
\label{gen_frame}

\subsection{The bAbI Question Answering Tasks}

The bAbI dataset is a series of 20 question answering tasks, comprising a unique training set and test set for each task. Within each set, the data represents a series of text sentences that describe a simple contextual domain, which we refer to as ``bAbI-world''. Both the training and test sets are divided into discrete sequences of statement sentences that form a particular story. Each story is either terminated or interspersed with questions that refer to either the current state or some historical condition of the contemporary story domain.

The bAbI dataset was presented in \citet{weston2015towards} and has provided the basis for a variety of recent attempts to develop systems that are able to satisfy complex reasoning tasks \citep{weston2015towards,weston2014memory, sukhbaatar2015end,sukhbaatar2015weakly,kumar2015ask}. The bAbI dataset, with background research and the accompanying test methodology is available at \url{https://research.facebook.com/researchers/1543934539189348}.

\subsection{Dynamic Adaptive Network Intelligence (DANI)}

Dynamic Adaptive Network Intelligence (DANI) is a general method for learning a dynamic structural representation of the world, based on the strength of contextual connections perceived between discrete data over time. 

DANI is a graphical system that derives a learned representation of data using similarity distance measures between individual components of the global model and class partitions. In common with \citet{pujara2013ontology}, we find that the Jaccard similarity distance measure works well when modelling sparse, high-dimensional data, although a range of alternative distance measures can also be implemented successfully within the same model structure (for a detailed survey see \citet{choi2010survey}).

The key characteristics of DANI are the network representation of unique data and the continuous adaptation of the model space and evaluation parameters in response to new data. DANI can be implemented fully unsupervised (with the domain representation learned heuristically over time), with weak supervision, or fully supervised. In the case of full supervision; the system may refer to a secondary ontology or vocabulary, or prior weightings can be employed to accelerate learning. In this paper, we report the supervised learning of DANI, for bAbI tasks 1 to 20, and the application of the model with weak supervision to task 19.

\subsection{Data Representation}

The foundation of DANI is the learned representation of the relationships perceived between data instances. The primary measure of the strength of those relationships is a weighted similarity distance measure, in the form of a Jaccard coefficient \citep{jaccard1901etude}. Each time a relationship is perceived, the similarity measure between instances is updated. As new relationships between data instances are perceived, these contextual relationships are added to DANI.

For supervised learning of each bAbI task; we construct an undirected graphical model of the type that has become popular through its application to problems in a diverse range of fields, including; communications engineering, social network analysis, neuroscience, biology, and geophysical studies \citep{denli2014multi}. For each story $S$ within the bAbI training set, the bAbI domain model $\hat{B}$ comprises a simple undirected graph, $B$, defined by: (i) a nonempty set of vertices 
$V(B)$; (ii) a set of connecting edges $E(B)$; (iii) a set of annotated vertex attributes $\Theta (V(B)) \subseteq \Theta$; and (iv) a set of annotated edge attributes $\Theta (E(B)) \subseteq \Theta$.  Hence;  $B = \{V, E, \Theta \}$ and $S_{t} \equiv B_{t}$, the contemporary learned representation of the domain.

For weakly supervised learning of bAbI task 19, we define an independent graphical model $\hat{M}$ that learns the contextual association of the task attributes. The graph $M$ is populated by the edge attributes $\Theta (E(B)) \subseteq \Theta$ and a null class: ``unknown'' $\in\Theta$, where the null class represents as yet unobserved domain attributes. The secondary model, $\hat{M}$, evaluates the mutual association of contextual attributes within the bAbI sentences (e.g. ``north'', ``south'', ``east''and ``west'') as they are observed by the system. Where new attributes are observed they are appended to the set of graph vertices $V(M)$; and correlated by an undirected graph edge.

In common with both Ising models and Hopfield networks \citep{murphy2012machine}, we apply a symmetrical weighting of the type $w_{\alpha\beta}\leftrightarrow w_{\beta\alpha}$ as a model attribute $\Theta(w_{\alpha\beta})$ on the unique edge connecting vertices $v_{\alpha}$ and $v_{\beta}$ of the learner $\hat{M}$.  The model attribute $\Theta(w_{\alpha\beta})$ reflects a dynamic measure of the binary vector symmetry observed between unique entity pairs $(\alpha,\beta)$ over the set of data samples, $S$.  We compute the symmetry measure by reference to binary variables whose simultaneous relationship is codified by a $2\times2$ contingency table of the type shown in Table~\ref{maths1}.  

To accommodate the sparse, high-dimensional, nature of the application dataset, we neglect the absence of entity pair combinations and focus only on their coincidence within the training dataset. This approach is consistent with a variety of binary similarity coefficients \citep{warrens2008similarity,choi2010survey} that do not operationalise incidences of mutual absence (represented in Table~\ref{maths1} by quantity $d$). We find that the Jaccard similarity coefficient, $sJac$, provides a robust measure over the suite of bAbI tasks, where:

\begin{equation}
sJac =\frac{a}{b+c-a}
\end{equation}

The Jaccard coefficient $\Theta(w_{\alpha\beta})$ was continuously updated for all task attribute relations and the null class during the task training cycle.

\begin{table}[ht]
\caption{$2\times2$ contingency table for unique dimensions $x_\alpha$ and $x_\beta$ of the bAbI-world domain, $B$, that occur independently in stories, $s$.}
\label{maths1}
\begin{center}
\begin{tabular}{ |P{3cm}|P{3cm}|P{3cm}|}
 \hline
 $\forall s$   &$|s\mapsto x_{\beta|_\beta} =1$   & $|s\mapsto x_{\beta|_\beta} =0$ \\
 \hline
 $|s\mapsto x_{\alpha}|_{\beta} =1$       &$a$  &$b$\\
 \hline
 $|s\mapsto x_{\alpha}|_{\beta} =0$        &$c$ &$d$ \\
 \hline
\end{tabular}
\end{center}
\end{table}

The DANI model architecture employed by this study is shown in Figure \ref{fig:text1}:

\begin{figure}[ht]
\begin{center}
\includegraphics[width=140mm,scale=0.5]{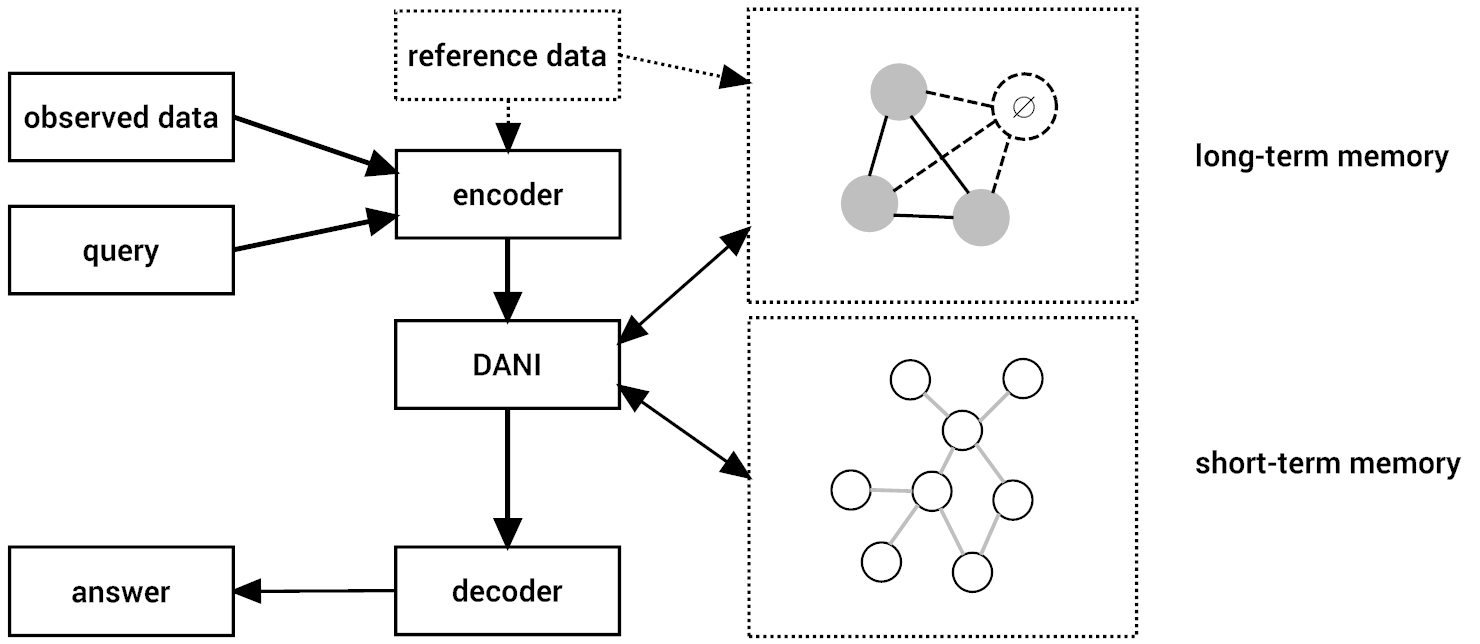}
\end{center}
\caption{The DANI model architecture.}
\label{fig:text1}
\end{figure}

The finite nature of the bAbI task datasets is easily accommodated by the DANI model architecture. For real-world applications, with complex class associations and contextual expansion, we have successfully implemented variants of DANI featuring dynamic normalization and compression of the model space. For the reported study, no parametric restriction of the model space was required.

\subsection{Training DANI}

For the bAbI question answering tasks, we tested two versions of DANI to demonstrate comparative results. The fully supervised version (DANI-S) was initialised with logical primitives (e.g. the word ``the'' prefixes a bAbI-world entity such as ``bathroom'') and a vocabulary of operational terms drawn from the training dataset. According to \citet{weston2015towards}, there were 150 words in the training set.  These were assigned to broad classes such as; ``name'', ``item'', ``shape'', ``colour'', ``place'', etc. 

DANI-S used a set of logical primitives to decode and represent the statements given in the training data. Decoding of the semantic statements was undertaken by reference to a basic set of definitive grammatical terms; e.g. ``to'',``from'', ``of'', and ``the''. 

The second version of our system (DANI-WS) is weakly supervised, which means that DANI was only initialised with the definitive grammatical terms stated above and did not refer to an associated task vocabulary. The model efficiently learned the classes of entities, the task attributes and the structural relationships encoded within the training data by observing a minimum number of training examples $(n=18)$.

During the training cycle, the ``bAbI-world''graph $B$ was constructed for each story and the model graph $M$ was revised to reflect correspondence between the answer to each training question and the knowledge of the story encoded in $B$. At the end of each story, the ``bAbI-world'' graph was cleared, while the model graph $M$ was persisted until the end of training.

At test time, the learned data representation forming graph $M$ was fixed and all training parameters were cleared. We did not explicitly indicate the supporting sentences within the text during training or test for either version of DANI.  The answer was not available to the system at test time and answers were derived only from the learned state of the model representation $\hat{M}$.

\subsection{Information Retrieval}

The bAbI tasks that have proved the most challenging to solve in a weakly supervised manner in \citet{weston2015towards,weston2014memory, sukhbaatar2015end,sukhbaatar2015weakly,kumar2015ask} appear to fall into two categories.  Tasks 2, 3, 4, 5, 7 and 8 all require the model to identify relevant data and hold it in an episodic memory over a chain of events.  Tasks 16, 17, 18 and 19 require reasoning over implicit relationships in the data. 

Modelling the story data as an undirected network $B$ acts as a short-term memory for both training and test purposes. In other applications, we have introduced gradual emergence and decay of the veracity of entities within the domain space to maintain the temporal relevance of the domain model where it is not periodically cleared.  The attribute model $\hat{M}$ that is trained over the training data constitutes a long-term system memory. Although the long-term memory component of DANI is fixed at test time for the reported application; in practice, the long-term memory can continue to learn the task context indefinitely. 

During the test cycle, answers are either directly derived from short-term memory of the current story $\hat{B}$, or are generated from long-term memory $\hat{M}$ by evaluation of the similarity distance measures between complementary attributes. Answers are extracted from short-term memory by means of a simple path query and evaluation of the attributes on the path edges. A visual example of how the answer to task 19 can be queried within the network is provided in Figure \ref{fig:text3} in the Appendix to this paper.

In the case of DANI-WS, the answers that are generated for the bAbI tasks from long-term memory $\hat{M}$ may be ambiguous, due to the small number of classes evident in the data. Selection of unique answers is made through continuous refinement of potential answers by examining the diminishing similarity of the answer candidate and the candidate path with rejected possibilities. This cycle of candidate evaluation is repeated to exhaustion, which is indicated by the null attribute --- since the null attribute is the condition for which an answer is sought. Integration of the null attribute as a candidate class was found to resolve ambiguity without exhaustion of potential paths.

The incorporation of a null class within long-term memory $\hat{M}$ is important, as it represents as yet unknown elements of the task domain. The inclusion of unknown information helps to control for spurious association of known entities, while reinforcing the factual relationships between entities that are observed by their mutual coincidence in the task answers. Over time, this methodology forms a powerful and accurate basis for precise inference based on the dynamic properties of the data.

\section{Related Work}
\label{related}

Recent research by \citet{berant2013semantic,berant2014modeling} on graphical approaches to reading comprehension focused on methods to map question-answer pairs into a static knowledge graph (KG) of facts using latent logical forms.  They then use the KG as a memory for information retrieval.  Rather than seeking to map from phrases to logical predicates, DANI seeks to learn using only those logical primitives required for feature extraction from the text input. The learned data representation provides a contextual mapping that is independent of the data observed, which is why it is an approach that can be applied to multiple domains.  

\citet{hixonlearning}(2015) recently described a new system that learns a KG from open, natural language dialogs using task-driven relations.  In common with the approach to learning adopted by \citet{hixonlearning}(2015); rather than seeking to build a fixed KG of a domain or reference to prescribed ontological rules \citep{pujara2013ontology}, DANI learns a polymorphic KG continuously, using the similarity distance between the concepts that have been observed.

\citet{gu2015traversing} recently demonstrated that performing path queries on a knowledge graph is an efficient means of inferring missing information for question answering tasks.  When seeking to make inferences over the KG, DANI conducts information retrieval over the structure of ``bAbI-world'' graph $B$. 

\section{Results}

\subsection{Baselines}

We compare our approach to the following recent research:

\begin{itemize}  

        \item Strongly supervised AM+NG+NL Memory Networks (MemNN) proposed in \citet{weston2015towards}.  

        \item A standard weakly supervised Long Short Term Memory (LSTM) model as reported in \citet{weston2015towards}.  

        \item Weakly Supervised End-to-End Memory Networks (MemN2N) proposed in \citep{sukhbaatar2015end}.

        \item DANI-S: Fully supervised vocabulary-based model

        \item DANI-WS: Weakly supervised model with logical primitives.

 		\end{itemize}

\subsection{Results}

Shallow learning systems have historically been limited in the complexity of the functions they could compute because of the difficulty of extracting a suitable vector of features. DANI is able to overcome this by conditioning the model space by compression and structural reconfiguration, while retaining definite contextual association of sparse data. 

Continuous adaptation of the network model, in correspondence with the changing similarity measures, maintains the computational efficiency of the system when scaling. In separate applications of our system, we have found that the DANI model architecture exhibits consistent performance that approximates to the polylogarithmic class $O(\log{}n/2)^{2}$, where $n \propto |V(G)|$ (in the case of this study ``bAbI-world'' graph $B=G$).

The reported results for our system were obtained using a commercially available laptop computer without the use of parallel computing or any other form of system optimization. The host computer included: a 2.8GHz Dual-core Intel® Core™ i7 processor; a 16GB 1600MHz DDR3L SDRAM memory module; and, a 512GB flash storage device that was used to store the sample dataset and the persisted undirected graph model. The training time was negligible, being of the order of a single second for both DANI-S and DANI-WS, over the 1,000 sample training sets. The test time for each task was also trivial, with each test of 1,000 questions being completed in the order of a second. 

We applied the supervised version of our system, DANI-S, to the full suite of bAbI question answering tasks, numbers 1 to 20. To enable early presentation of our results; we applied the weakly supervised version, DANI-WS, to task 19 only, as this task has proved the most difficult for comparable systems. We plan to extend the application of DANI-WS to the remaining bAbI tasks and to update the preliminary results reported in the paper, in due course.

Our test methodology adhered to the guidance provided by the authors of the bAbI dataset. No specific engineering of the DANI system was undertaken in order to accommodate the test requirements and data structures of the individual tasks. Test results for the DANI architecture, with comparison to those of previous studies, are reported in Table~\ref{sample-table} and Table~\ref{sample-table2}.

The supervised version of the model, DANI-S, does produce a small number of errors at test time. These are mainly due to duplication issues within the model, for example in task 3 and 14, if an item was handled by a number of different people in the same room or a person moved back to a room they had visited previously.  Rather than engineering these errors out by discriminating between multiple instances, we wanted to keep the query function as simple as possible in order to demonstrate that the bulk of the questions can be answered within a single DANI model structure.

\begin{table}[h]
\caption{Strongly supervised test error rates (\%) for DANI-S.  Compared with AM+NG+NL MemoryNetworks (MemNN) as presented in \citet{weston2015towards}.}
\label{sample-table}
\begin{center}
\begin{tabular}{lrr}
\\
\multicolumn{1}{c}{\bf Task}  &\multicolumn{1}{c}{\bf MemNN} &\multicolumn{1}{c}{\bf DANI-S}
\\ \hline \\
1 - Single Supporting Fact         &0.0 	&0.0 \\
2 - Two Supporting Facts           &0.0		&0.0 \\
3 - Three Supporting Facts         &0.0		&4.2 \\
4 - Two Arg. Relations             &0.0		&0.0 \\
5 - Three Arg. Relations           &2.0		&0.4 \\
6 - Yes/No Questions               &0.0		&0.0 \\
7 - Counting           			   &15.0	&0.0 \\
8 - Lists/Sets           		   &9.0		&0.0 \\
9 - Simple Negation            	   &0.0		&0.0 \\
10 - Indefinite Knowledge          &2.0		&0.0 \\
11 - Basic Coreference             &0.0		&0.0 \\
12 - Conjunction           		   &0.0		&0.0 \\
13 - Compound Coreference          &0.0		&0.0 \\
14 - Time Reasoning           	   &1.0		&3.6 \\
15 - Basic Deduction           	   &0.0		&0.0 \\
16 - Basic Induction           	   &0.0		&3.6 \\
17 - Positional Reasoning          &35.0	&0.0 \\
18 - Size Reasoning                &5.0		&0.7 \\
19 - Path Finding            	   &64.0	&0.0 \\
20 - Agent's Motivations           &0.0		&0.0 \\
\hline 
Mean Error         	   &6.7		&0.6 \\
\end{tabular}
\end{center}
\end{table}

\begin{table}[ht]
\caption{Weakly supervised test error rates (\%) for Task 19 - Path Finding for DANI-WS.  Compared with: \citep{weston2015towards} reported results for a Long-Short Term Memory (LSTM) architecture. \citep{sukhbaatar2015end} reported results for various designs for End-to-End Memory Networks including bag-of-words representation (BoW), position encoding representation (PE), linear start training (LS) and random noise injection of time index noise (RN).}
\label{sample-table2}
\begin{center}
\begin{tabular}{ |P{1.3cm}|P{1.3cm}|P{1.3cm}|P{1.3cm}| P{1.3cm}|P{1.3cm}||P{1.3cm}|P{1.3cm}|  }
 \hline
 \multicolumn{6}{|c||}{Baseline} & \multicolumn{2}{|c|}{DANI}\\
 \hline
 LSTM   & MemN2N BoW    & MemN2N PE &   MemN2N PE LS & MemN2N PE LS RNN & No. of training ex.   & DANI-WS  & No. of training ex.\\
 \hline
 92.0         &89.7 &87.4  &85.6 &82.8 &1,000 &0.0 &18\\

 \hline
\end{tabular}
\end{center}
\end{table}

In our application of DANI-WS to task 19, during training we restricted the update procedure of the long-term memory $\hat{M}$ so that it was only revised after the observation of model answers. In task 19, model answers are provided in training after 5 prior statements, with blocks of 6 sample sentences forming a discrete story. We found that only 3 stories ($n=18$ sample sentences) were required to train the long-term memory $\hat{M}$ with sufficient accuracy to achieve the reported test results. Training over additional sentences improved the intrinsic validity of the learning representation, but a 0\% error rate was already attained. Tuning of the compression algorithm may enable training over a single story (since the second story in the training data for task 19 does not impart new information to the learner). However, we consider that tuning the DANI-WS model would constitute specific engineering of the system to improve performance.  We plan to expand our experiment to evaluate the threshold for accurate learning representation of the remaining 19 tasks.

\section{Conclusions}
\label{conclusions}

In this paper, we described a new approach to learning an adaptive network-based representational model of the bAbI dataset, based on the contextual similarity between concepts. We tested the performance of this system against the question-answering tasks and reported state-of-the-art results. Our preliminary results for the supervised version of the model indicate that this model architecture is an efficient approach solving to a variety of question-answering tasks with low computational overhead. The preliminary results reported for the weakly supervised model, DANI-WS, suggest that this version of our system has the potential to efficiently generalise with significantly less supervision or computational complexity than other contemporary approaches.

Although we report the application of DANI as an independent framework for learning representation, we recognize that our system could be employed to condition the input and intermediate layers of neural architectures of the type reported in comparable studies \citep{weston2015towards,weston2014memory,sukhbaatar2015end,sukhbaatar2015weakly,kumar2015ask}. We intend to explore this opportunity and to update our preliminary results during the next phase of our research.

\bibliography{iclr2016_conference}
\bibliographystyle{iclr2016_conference}

\appendix
\section{Visual example of task 19}

\begin{figure}[ht]
\begin{center}
\includegraphics[width=70mm,scale=0.1]{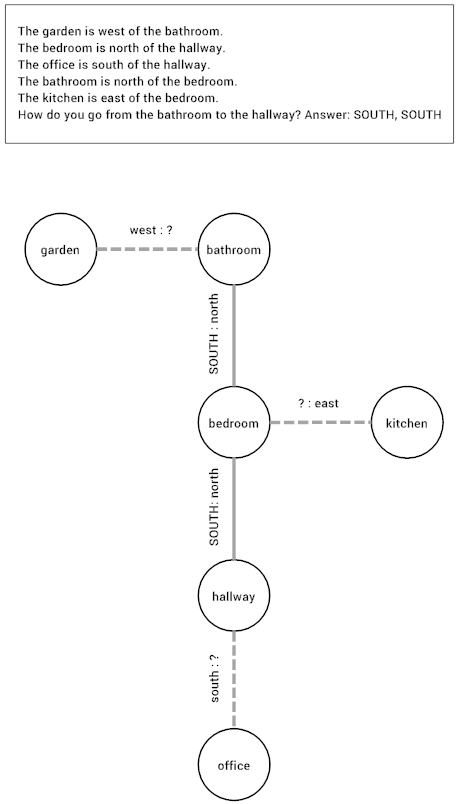}
\end{center}
\caption{Visual example of how the answer to task 19 can be queried within ``bAbI-world'' graph $B$.}
\label{fig:text3}
\end{figure}

\end{document}